\documentclass{article}

\usepackage{arxiv}

\usepackage[utf8]{inputenc} 
\usepackage[T1]{fontenc}    
\usepackage{hyperref}       
\usepackage{url}            
\usepackage{booktabs}       
\usepackage{amsfonts}       
\usepackage{nicefrac}       
\usepackage{microtype}      
\usepackage{lipsum}

\usepackage{cite,comment}
\usepackage{amsmath,amssymb,amsfonts}
\usepackage{graphicx}
\usepackage{textcomp}
\usepackage{xcolor}
\usepackage{xspace}
\usepackage{tikz}
\usepackage{pgfplots}
\usetikzlibrary{arrows,backgrounds,decorations,decorations.pathmorphing,positioning,fit,automata,shapes,snakes,patterns,plotmarks,calc}
\usepackage[vlined,ruled,linesnumbered]{algorithm2e}
\usepackage{stmaryrd}
\usepackage[inline]{enumitem}
\usepackage{booktabs}
\usepackage{nth}

\graphicspath{{Fig/}}

\usepackage{amsthm}

\usepackage{graphicx}
\usepackage{caption,subcaption}
\usepackage{bm}
\usepackage{amsmath}

\DontPrintSemicolon

\newcommand{\Reals}{\mathbb{R}}

\newcommand{\mypara}[1]{\vspace{0.3em} \noindent {\bf #1.\ }}
\newcommand{\myipara}[1]{\vspace{0.3em} \noindent {\em #1.\ }}

\newcommand{\DiffRNN}{{\scshape DiffRNN}}
\newcommand{\ReluDiff}{{\scshape ReluDiff}}

\newcommand{\Crown}{{\scshape Crown}}
\newcommand{\Popqorn}{{\scshape Popqorn}}

\newcommand{\inputspace}{X}
\newcommand{\outputSpace}{Y}

\newcommand{\inputDim}{m}

\newcommand{\Nat}{\mathbb{N}}

\newcommand{\vInputSeq}{\mathbf{x}}
\newcommand{\vOutputSeq}{\mathbf{y}}

\newcommand{\outputDim}{p}

\newcommand{\numNeurons}{\ell}
\newcommand{\vHid}{\mathbf{h}}

\newcommand{\whh}{W_{\vHid\vHid}}
\newcommand{\whx}{W_{\vHid\vInputSeq}}
\newcommand{\whout}{W_{\vHid\vOutputSeq}}
\newcommand{\hbias}{\mathbf{b}_{\vHid}}
\newcommand{\biasout}{\mathbf{b}_{\vOutputSeq}}

\newcommand{\act}{\mathbf{\sigma}}

\newcommand{\vAff}{\mathbf{a}}

\newcommand{\ignore}[1]{}

\newcommand{\ReLU}{\emph{ReLU}}
\newcommand{\Sigmoid}{\emph{Sigmoid}}

\newcommand{\sigmoid}{\mathit{\sigma_\mathcal{S}}}

\newcommand{\diffa}{\bm{\delta}^{\vAff}}
\newcommand{\diffh}{\bm{\delta}^{\vHid}}
\newcommand{\diffy}{\bm{\delta}^{\vOutputSeq}}

\newcommand{\igate}{\mathbf{i}}
\newcommand{\fgate}{\mathbf{f}}
\newcommand{\ggate}{\mathbf{g}}
\newcommand{\ogate}{\mathbf{o}}
\newcommand{\cstate}{\mathbf{c}}

\newcommand{\wih}{W_{\igate\vHid}}
\newcommand{\wix}{W_{\igate\vInputSeq}}
\newcommand{\wvx}{W_{\bm{v}\vInputSeq}}
\newcommand{\ibias}{\mathbf{b}_{\igate}}

\newcommand{\wfh}{W_{\fgate\vHid}}
\newcommand{\wvh}{W_{\bm{v}\vHid}}
\newcommand{\wfx}{W_{\fgate\vInputSeq}}
\newcommand{\fbias}{\mathbf{b}_{\fgate}}

\newcommand{\wgh}{W_{\ggate\vHid}}
\newcommand{\wgx}{W_{\ggate\vInputSeq}}
\newcommand{\gbias}{\mathbf{b}_{\ggate}}

\newcommand{\woh}{W_{\ogate\vHid}}
\newcommand{\wox}{W_{\ogate\vInputSeq}}
\newcommand{\obias}{\mathbf{b}_{\ogate}}

\newcommand{\diff}{\bm{\delta}}
\newcommand{\diffi}{\bm{\delta}^{\igate}}
\newcommand{\difff}{\bm{\delta}^{\fgate}}
\newcommand{\diffg}{\bm{\delta}^{\ggate}}
\newcommand{\diffo}{\bm{\delta}^{\ogate}}
\newcommand{\diffc}{\bm{\delta}^{\cstate}}
\newcommand{\diffv}{\bm{\delta}^{\bm{v}}}

\newcommand{\uin}{\bm{in}}

\title{DiffRNN: Differential Verification of Recurrent Neural Networks}

\author{
  Sara Mohammadinejad\\
  University of Southern California\\
  \texttt{saramoha@usc.edu} \\
     \And
  Brandon Paulsen\\
  University of Southern California\\
  \texttt{bpaulsen@usc.edu} \\
     \And
  Chao Wang\\
  University of Southern California\\
  \texttt{wang626@usc.edu} \\
     \And
  Jyotirmoy V. Deshmukh\\
  University of Southern California\\
  \texttt{jdeshmuk@usc.edu} \\
}
\begin{document}
\maketitle

\begin{abstract}
Recurrent neural networks (RNNs) such as Long Short Term Memory (LSTM)
networks have become popular in a variety of applications such as
image processing, data classification, speech recognition, and as
controllers in autonomous systems. In practical settings, there is
often a need to deploy such RNNs on resource-constrained platforms
such as mobile phones or embedded devices. As the memory footprint and
energy consumption of such components become a bottleneck, there is 
interest in compressing and optimizing such networks using a range of
heuristic techniques.  However, these techniques do not guarantee the
safety of the optimized network, e.g., against adversarial inputs, or
equivalence of the optimized and original networks.
To address this problem, we propose \DiffRNN{}, the first differential
verification method for RNNs to certify the equivalence of two
structurally similar neural networks. 
Existing work on differential verification for \ReLU{}-based
feed-forward neural networks does not apply to RNNs where nonlinear
activation functions such as \Sigmoid{} and \emph{Tanh} cannot be avoided.
RNNs also pose unique challenges such as handling sequential inputs,
complex feedback structures, and interactions between the gates and
states.
In \DiffRNN{}, we overcome these challenges by bounding nonlinear
activation functions with linear constraints and then solving
constrained optimization problems to compute tight bounding boxes on
non-linear surfaces in a high-dimensional space.  The soundness of
these bounding boxes is then proved using the \emph{dReal} SMT
solver. We demonstrate the practical efficacy of our technique on a variety 
of benchmarks and show that \DiffRNN{} outperforms state-of-the-art
RNN verification tools such as \Popqorn{}.


\end{abstract}

\section{Introduction}
Deep neural networks, and in particular, recurrent neural networks
(RNNs), have been successfully used in a wide range of applications
including image classification, speech recognition, and natural
language processing. However, their rapid growth in safety-critical
applications such as autonomous driving~\cite{bojarski2016end} and
aircraft collision avoidance~\cite{julian2019deep} is accompanied by
safety concerns \cite{lyu2019fastened}. For example, neural
networks are known to be vulnerable to adversarial
inputs~\cite{GoodfellowSS15,szegedy2013intriguing}, which are security
exploits designed to fool the neural
networks \cite{KurakinGB17a,Moosavi-Dezfooli16,NguyenYC15,XuQE16}.

In addition, trained neural networks typically go through changes before deployment,
thus raising concerns that the changes may introduce new behaviors.
Specifically, since neural networks are computationally and memory
intense, they are difficult to deploy on resource-constrained devices~\cite{HanMD16,cheng2017survey}. 
Network compression
techniques (such as edge pruning, weight quantization, and neuron
removal) are often needed to reduce the network's
size~\cite{cheng2017survey}. Compression techniques typically
use mean-squared error over sampled inputs as a performance measure 
to test equivalence. Such a measure is statistical, and does 
not provide formal worst-case guarantees on the deviation between behaviors of two networks.

While there are recent efforts on applying differential
testing~\cite{ma2018deepgauge,PeiCYJ17,TianPJR18} and
fuzzing~\cite{odena2018tensorfuzz,xie2019deephunter,xie2019diffchaser}
techniques to neural networks, they can only increase the confidence
that the networks behave as expected for some of the inputs.  However,
they cannot prove the equivalence of the networks for all inputs.
To the best of our knowledge, \ReluDiff{}~\cite{paulsen2020reludiff} is the
only tool that aims to prove the equivalence of two neural networks
for all inputs.  \ReluDiff{} takes as input two feed-forward neural 
networks with piecewise linear activation functions known as rectified 
linear units ( \emph{ReLU}). The \emph{ReLU} activation essentially 
allows the neural network to be treated as a piecewise linear (PWL) 
function (with possibly many facets/pieces).

\ReluDiff{} exploits the PWL nature of activations, and hence cannot 
	natively handle non-PWL activation functions like \emph{Sigmoid},
	\emph{Tanh}, and \emph{ELU}, let alone the more complex operations
	of LSTMs, which take the \emph{product} of these non-linear functions,
	e.g. \emph{Sigmod}$ \times $\emph{Tanh}. This poses significant
	limitations because popular libraries ``hardcode''
	\emph{Tanh} and \emph{Sigmoid} for some, or all of the activation
	functions in the network. For example, Fig.~\ref{fig:lstm} shows the LSTM
	structure hardcoded into Tensorflow. Thus, for RNNs, we need a technique that
	can handle these challenging and arbitrary nonlinearities. 
	In addition, we face several other unique challenges when considering 
	RNNs, including how to \emph{soundly and efficiently} handle (1) sequential 
	inputs, (2) the complex feedback structures, and (3) interactions
	between the gates and states.

To overcome these challenges, we propose \DiffRNN{}, the first
differential verification technique for
bounding the difference of two structurally similar RNNs.
Formally, given two RNNs that only differ in numerical values of their edge weights, 
denoted $\vOutputSeq=f(\vInputSeq)$ and $\vOutputSeq^\prime =f^\prime(\vInputSeq)$, 
where $\vInputSeq\in X$
is an input, $X$ is an input region of interest, and $\vOutputSeq, \vOutputSeq^\prime$ are
the outputs, \DiffRNN{} aims to prove that
$\forall \vInputSeq \in X~.~ |f^\prime (\vInputSeq) - f(\vInputSeq)| < \epsilon$, where
$\epsilon$ is a reasonably small number. 

\begin{figure}
\centering
\includegraphics[scale=0.50]{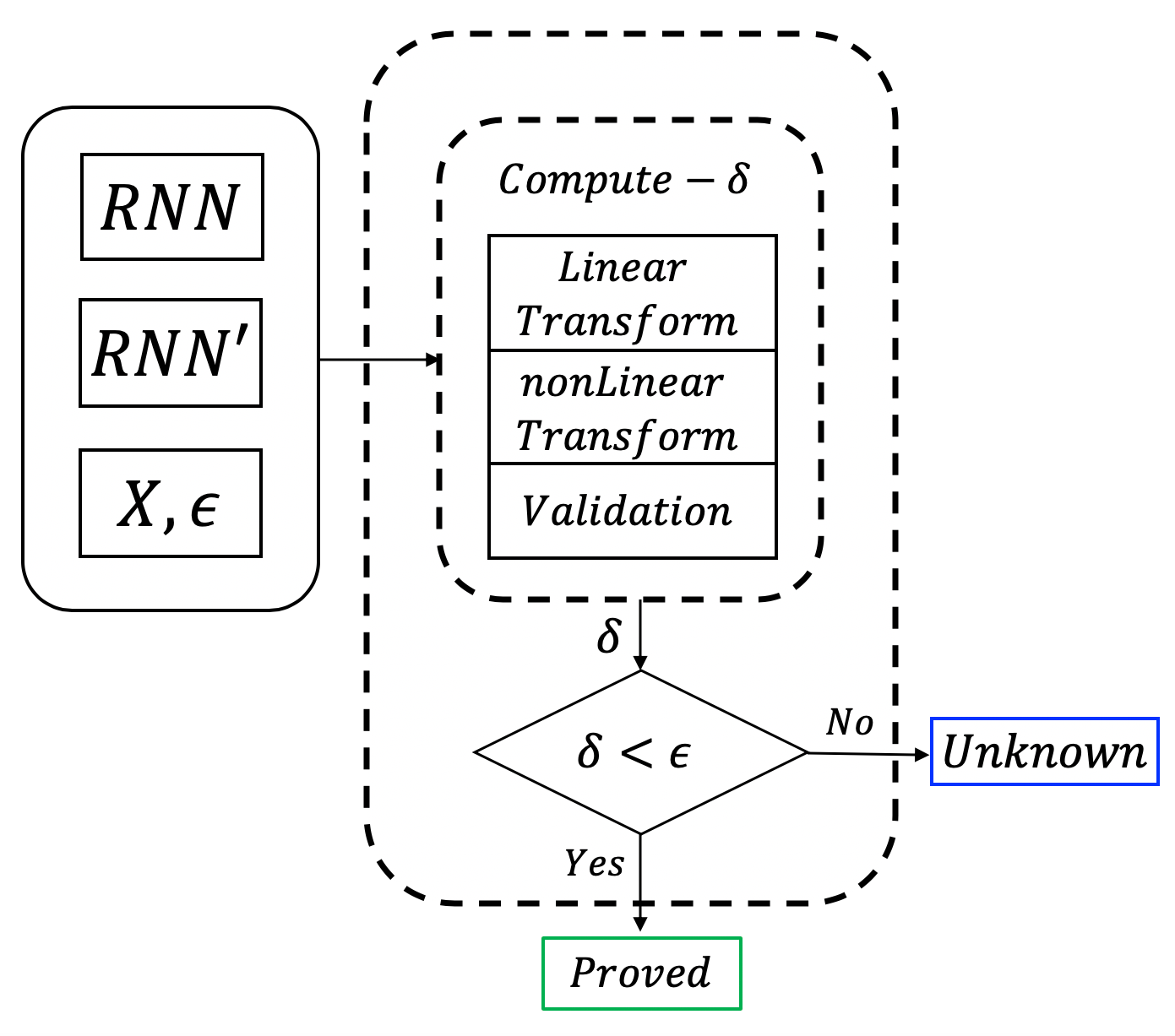}
\caption{The differential verification flow of \DiffRNN{}. $\delta$ is the difference interval, and $\epsilon$ is the bound on the output differences of the compressed and original networks.}
\label{fig:highlevel}
\end{figure}

Fig.~\ref{fig:highlevel} shows the high-level flow of \DiffRNN{},
whose input consists of two networks, $RNN$ and
$RNN^\prime$, an input region, $X$, and a small difference bound
$\epsilon$.  It produces two possible outcomes: \textit{Proved},
or \textit{Unknown}.
Internally, \DiffRNN{} uses symbolic interval arithmetic to compute linear bounds
on both the output values of each network's neurons and the
\emph{differences} between the neurons of the two networks.
We compute these linear bounds \emph{efficiently} in a layer-by-layer
fashion, that is, using the bounds of the previous layer to compute the
bounds of the current layer. If the bounds on the final output difference satisfy $ \epsilon $,
\DiffRNN{} returns \emph{Proved}, otherwise it returns \emph{Unknown}.

To compute the output difference \emph{accurately}, we bound nonlinear activation functions with
linear constraints and then solve constrained optimization problems to
obtain tight bounding boxes on nonlinear surfaces in a
high-dimensional space.  We also prove the soundness of these bounding
boxes using \emph{dReal}~\cite{gao2013dreal}, which is an off-the-shelf \emph{delta-sat} SMT solver\footnote{\emph{dReal} is implemented based on delta-complete decision procedures; it returns either unsat or delta-sat on the given input formulas, where delta is a user-defined error bound \cite{gao2013dreal}.} that
supports nonlinear constraints.

\begin{figure}
\centering
\includegraphics[scale=0.65]{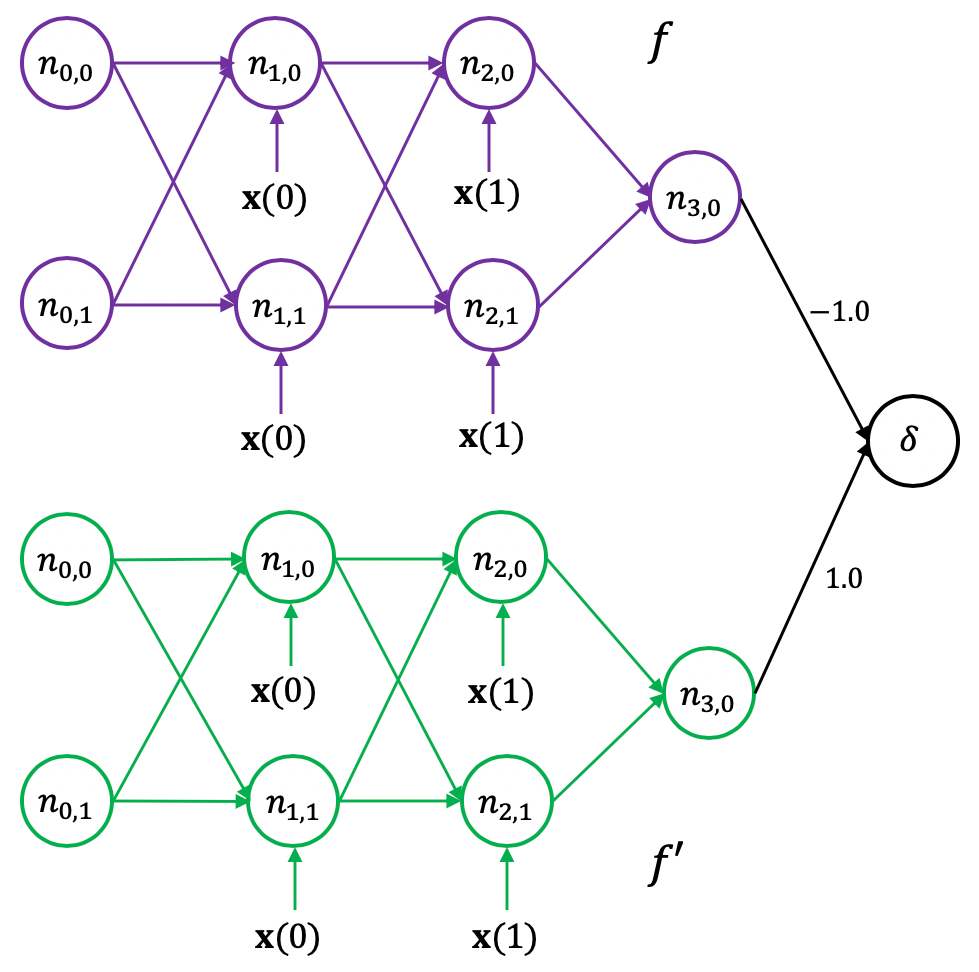}
\caption{Na\"ive differential verification of RNNs.}
\label{fig:rnn_diff_naive}
\end{figure}

While one could try and adapt an existing single-network
verification tool to solve our problem, in practice,
the bounds computed by this approach are too loose, since existing tools are
not designed to exploit the relationships between neurons in two RNNs.
To confirm this observation, we constructed the following experiment.
We took two \emph{identical} networks $f(\vInputSeq)$ and $ f'(\vInputSeq) $, i.e., with the same network topology and edge weights. We then constructed a new network
$ f''(\vInputSeq) = f'(\vInputSeq) - f(\vInputSeq) $, illustrated in Fig.~\ref{fig:rnn_diff_naive}.
Then, we took \Popqorn{}, a state-of-the-art RNN verification tool, and
attempted to prove $ f''(\vInputSeq) < \epsilon $ for all $ \vInputSeq $. While \Popqorn{} could
not prove this for any $ \epsilon < 2.0 $, \DiffRNN{} could prove it easily for any $ \epsilon > 0 $.

We have implemented our proposed method and 
evaluated it on a variety of differential verification tasks involving
networks for handwritten digit recognition (MNIST) \cite{lecun-mnisthandwrittendigit-2010} and human activity recognition \cite{anguita2013public}.
Our results show that \DiffRNN{} is efficient and effective in
certifying the functional equivalence of RNNs after
compression techniques are applied.
We also compared \DiffRNN{} with \Popqorn{}~\cite{ko2019popqorn}, the state-of-the-art RNN verification
tool.
Our results show that \DiffRNN{} significantly outperforms \Popqorn{}~\cite{ko2019popqorn}: On average \DiffRNN{} is 2.73X more accurate and $60\%$ faster.

To summarize, our main contributions  are as follows:
\begin{itemize}[leftmargin=1.2em,topsep=0pt,itemsep=0pt]
\item 
We propose \DiffRNN{}, a differential verification method for proving
the functional equivalence of two structurally similar RNNs.
\item 
We develop techniques to handle the recursive nature of RNNs and
nonlinear functions such as \Sigmoid{} and \emph{Tanh}.
\item 
We develop techniques to handle both Vanilla RNNs and the more complex
LSTMs.
\item 
We formally verify the soundness of our linear approximation
techniques using \emph{dReal} \cite{gao2013dreal}.
\item 
We experimentally demonstrate that our method
significantly outperforms the state-of-the-art techniques.
\end{itemize}

\section{Background}
In this section, we review the basics of recurrent neural networks (RNNs),
including Vanilla RNNs and LSTMs\footnote{Gated recurrent units (GRUs) are structurally very similar to LSTMs, and differential verification hurdles for GRUs are the same as LSTMs; thus, we omit explaining GRUs in this paper for brevity.}, and interval bound propagation
(IBP), a technique for bounding the network's output values for all
input values.

\subsection{Recurrent Neural Networks}
\subsubsection{Vanilla RNNs}
A vanilla recurrent neural network is a function that maps time-indexed
{\em input sequences} to {\em output sequences}. Let $\inputspace$ be
a compact subset of $\Reals^\inputDim$, where $m$ is the number of input values at each time step.
An input sequence $\vInputSeq$
is a function from time $\{0,1,\ldots,T\}$ to input space $\inputspace$, where $T \in
\Nat$, and $\vInputSeq(j) \in \inputspace$ denotes the $j^{th}$ entry
in the time-indexed input sequence. An output sequence is a similar function that maps to
an output space $\outputSpace \subseteq \Reals^{\outputDim}$.

The structure of a vanilla RNN is as follows.  It consists of a single
layer of $\numNeurons$ neurons, and its output $\vHid$ at time $t$ depends on
(a) the output $\vHid$ at time $t-1$, and (b) the input $\vInputSeq$ at time $t$, as shown
below:
\begin{eqnarray}
\label{eq:vanilla_rnn}
\vAff(t) & = & \whh\cdot \vHid(t-1) + \whx\cdot \vInputSeq(t) + \hbias  \\
\vHid(t) & = & \act(\vAff(t)) \\
\vOutputSeq(t) & = & \whout\cdot \vHid(t) + \biasout \label{eq:output}
\end{eqnarray}

\noindent Here, $\vAff(t)$ is an intermediate variable that we
introduce to represent the affine transformation of the current input $\vInputSeq(t)$ and
previous state $\vHid(t-1)$. The weight matrices $\whh$ and $\whx$ have 
dimensions $\numNeurons \times \numNeurons$ and
$\numNeurons \times \inputDim$ respectively. The bias term $\hbias$ is
an $\numNeurons \times 1$ matrix. $\act$ is the nonlinear
component-wise {\em activation function} from $\Reals^\numNeurons$ to
$\Reals^\numNeurons$. We assume that $\vHid(0)$ is a fixed initial
state of the RNN at time $0$; it is a vector of size
$\numNeurons \times 1$. Finally, the output of the RNN at time $t$,
$\vOutputSeq(t)$ is defined as a
linear transformation of $\vHid(t)$, using the weight matrix $\whout$ and bias term $\biasout$. 

Thus, each multiplication above is a matrix multiplication, and for
all time steps $t$, $\vHid(t)$ and $\vInputSeq(t)$ are $\numNeurons$- and
$\inputDim$-length vectors, respectively. 
The activation function $\act$ may be  the sigmoid
activation ($\sigmoid$) or the hyperbolic
tangent activation ($\tanh$)\footnote{For a scalar input $u$, $\sigmoid(u) =
\frac{e^{u}}{1+e^{u}}$, and $\tanh(u) =
\frac{e^{u}-e^{-u}}{e^{u}+e^{-u}}$.}.

In differential verification, there is a second RNN whose parameters 
are denoted by $\vAff'$, $\vHid'$, $\vOutputSeq'$, $\whh'$, $\whx'$, 
$\whout'$, $\hbias'$ and $\biasout'$ respectively. The two RNNs under 
comparison are {\em structurally similar}, i.e., they 
only differ in the values of the edge weights, and have
the same activation functions. 

We also introduce $\diffa$, $\diffh$, and $\diffy$ to represent the differences:
$\diffa(t)=\vAff'(t)-\vAff(t)$, 
$\diffh(t) = \vHid'(t)-\vHid(t)$, 
and $\diffy(t) = \vOutputSeq'(t)-\vOutputSeq(t)$.

A many-to-one vanilla RNN differs from the vanila RNN model shown
above in one small way. For an input sequence of length $T$, the
output is computed only at time $T$, i.e., the final output of the
network is defined as $\vOutputSeq(T)$ (see Fig.~\ref{fig:vanilla_rnn} in Appendix).

\subsubsection{LSTMs}
\begin{figure}
\centering
\includegraphics[scale=0.7]{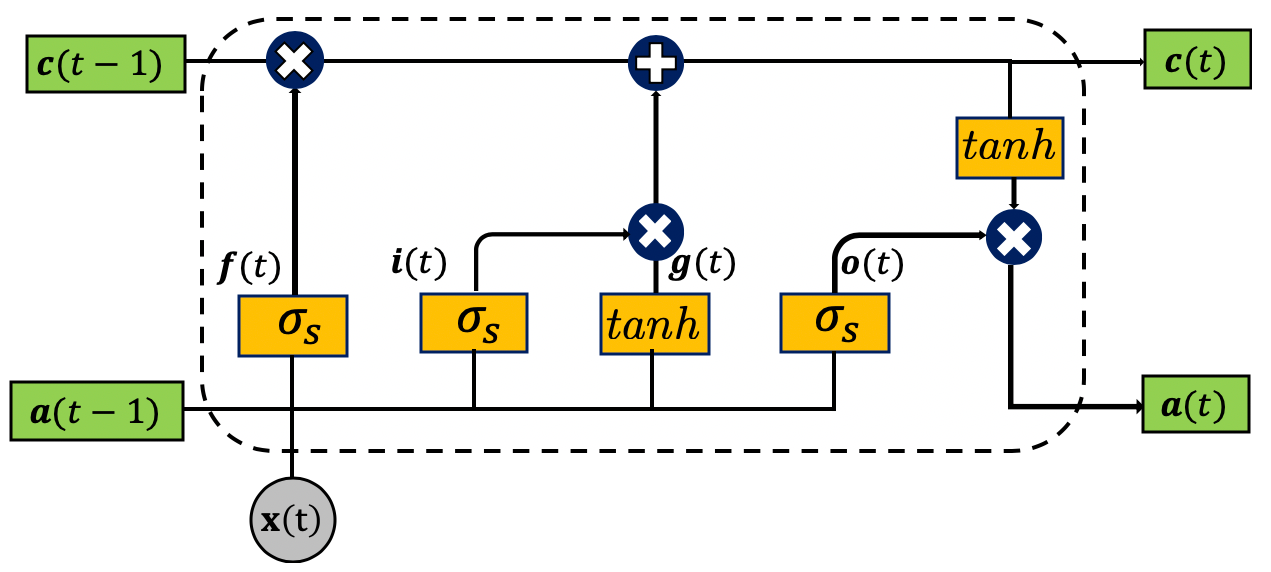}
\caption{An LSTM cell.}
\label{fig:lstm}
\end{figure}

Long short-term memory networks (LSTMs) were introduced to overcome
the limitation of Vanilla RNNs in learning long term sequential
dependencies \cite{sterin2017intrinsic}. Therefore, an LSTM is a
special kind of RNN, where each LSTM cell has four neurons that
interact with each other. As shown in Fig.~\ref{fig:lstm}, each LSTM
cell at time step $t$ takes $\cstate(t-1)$, $\vHid(t-1)$ and
$\vInputSeq(t)$ as input, and returns $\cstate(t)$ and $\vHid(t)$ as
output.  The input $\vInputSeq(t)$, the cell state $\cstate(t)$, and
the hidden state $\vHid(t)$ are all vectors of real values.
Thus, the four \emph{gates} and two \emph{states} within each LSTM
cell are evaluated as follows:

{\footnotesize
\begin{flalign*}
Input\;gate: \igate(t)&=\sigmoid(\wih \cdot  \mathbf{h}(t-1) +\wix\cdot \vInputSeq(t) +\ibias)\\
Forget\;gate: \fgate(t)&=\sigmoid(\wfh \cdot  \mathbf{h}(t-1) +\wfx\cdot \vInputSeq(t) +\fbias)\\
Cell\;gate:  \ggate(t)&=\tanh(\wgh \cdot  \mathbf{h}(t-1) +\wgx\cdot \vInputSeq(t) +\gbias)\\
Output\;gate:  \ogate(t)&=\sigmoid(\woh \cdot  \mathbf{h}(t-1) +\wox\cdot \vInputSeq(t) +\obias)\\
Cell\;state:  \cstate(t) &= \fgate(t) \odot \cstate(t-1)  + \igate(t) \odot \ggate(t)\\
Hidden\;state:  \vHid(t) &= \ogate(t) \odot \tanh(\cstate(t))\\
Output: \vOutputSeq(t) & = \whout\cdot \vHid(t) + \biasout
\end{flalign*}
}
%
Here, $\odot$ stands for Hadamard product (element-wise
multiplication).  Weight matrices $\wih$, $\wfh$, $\wgh$ and $\woh$
have dimensions $\numNeurons \times \numNeurons$. Weight matrices
$\wix$, $\wfx$, $\wgx$ and $\wox$ have dimensions $\numNeurons \times
\inputDim$.  Bias terms $\ibias$, $\fbias$, $\gbias$ and $\obias$ are
$\numNeurons \times 1$ matrices.
As before, $\sigmoid$ and $\tanh$ are the component-wise {\em
  activation functions} from $\Reals^\numNeurons$ to
$\Reals^\numNeurons$.  The input $\vInputSeq(t)$ is an
$\inputDim$-length vector, while $\igate(t)$, $\fgate(t)$,
$\ggate(t)$, $\ogate(t)$, $\cstate(t)$ and $\vHid(t)$ are all
$\numNeurons$-length vectors.

Similarly, we use $\igate'$, $\fgate'$, $\ggate'$, $\ogate'$,
$\cstate'$, $\vHid'$ and $\vOutputSeq'$ to represent parameters of the
second LSTM.  We also introduce the differences $\diffi(t)$,
$\difff(t)$, $\diffg(t)$, $\diffo(t)$, $\diffc(t)$ and $\diffh(t)$ as
vectors of size $\numNeurons \times 1$: For each $\mathbf{v} \in
{\igate,\fgate,\ggate,\ogate,\cstate,\vHid}$, we have
$\diff^\mathbf{v}(t) = \mathbf{v}'(t) - \mathbf{v}(t)$.

\subsection{Interval Bound Propagation (IBP)}

To soundly compute the output values of a neural network for all input
values, we represent these values as intervals, and use interval
arithmetic to compute their bounds.  

\subsubsection{Linear Operations}

Given two intervals, e.g., $p\in[a,b]$ and $q\in[c,d]$, the resulting
intervals of linear operations such as addition ($p+q$), subtraction
($p-q$), and scaling ($p \cdot c$, where $c$ is a constant) are well
defined.  That is,

{\footnotesize
\begin{align*}
[a, b] + [c, d] = [a+c, b+d]\\
[a, b] - [c, d] = [a-d, b-c]\\
[a, b] \cdot c = 
\begin{cases}
[a \cdot c, b \cdot c], & c\geq 0 \\
[b \cdot c, a \cdot c], & c<0
\end{cases}
\end{align*}
}

While the results are sound over-approximations, they may be overly
conservative.  For example, when $p=5x$, $q=4x$ and $x\in [-1,1]$,
since $p-q =5x-4x = x$, we know that $(p-q) \in [-1,1]$, but interval
subtraction returns $(p-q) = [-5,5] - [-4,4] = [-5-4,5-(-4)] =
[-9,9]$.  

A technique for improving accuracy is the use of symbolic inputs.
For example, instead of
using the concrete intervals $p\in[-5,5]$ and $q\in[-4,4]$, we may use 
the symbolic upper and lower bounds $p\in[5x,5x]$ and
$q\in[4x,4x]$, leading to $(p-q)=[5x-4x,5x-4x] = [x,x]$.  As a
result, we have $(p-q)= [-1,1]$ after concertizing the symbolic
bounds. 

In this work, we represent the \emph{symbolic} lower and upper bounds of $p$ as $L(p)$ and $U(p)$, and 
the concrete lower and upper bounds as $\underline{p}$ and $\overline{p}$, respectively.


%

\subsubsection{Non-linear Operations}

Sound intervals may also be defined for outputs of \Sigmoid{}
($\sigmoid$) and \emph{Tanh} ($\tanh$) activation functions.
Since both functions are \emph{monotonically increasing}, given a concrete 
input interval $p=[a,b]$, we have $\sigma(p) \in
[\sigma(a),\sigma(b)]$.
However, for a symbolic input interval, soundly approximating the
output is challenging. In an existing verification tool named
\Crown{}~\cite{zhang2018efficient}, e.g., this is solved by computing
linear bounds on the output of each activation function.
For LSTMs, the problem is even more challenging because it involves the \emph{product}
of nonlinear operations, such as $z = \sigmoid(x) \cdot \tanh(y)$ and $z= x \cdot
\sigmoid(y)$.  In
\Popqorn{}~\cite{ko2019popqorn}, for example, the output is bounded by
searching for linear bounding planes of the form $\alpha x + \beta y +
\gamma$, where $\alpha,\beta$ and $\gamma$ are computed using gradient
descent.

In this work, we build upon techniques from \Crown{} and \Popqorn{}
for bounding the output \emph{values} of network $f$'s neurons, to solve
the new problem of {\em bounding the differences} between two networks
$f$ and $f'$. 

\section{Overview}
In this section, we use an example to illustrate the high-level idea
of our method and the shortcomings of state-of-the-art single-network verification techniques for differential verification.

\begin{figure}
\centering
\includegraphics[scale=0.7]{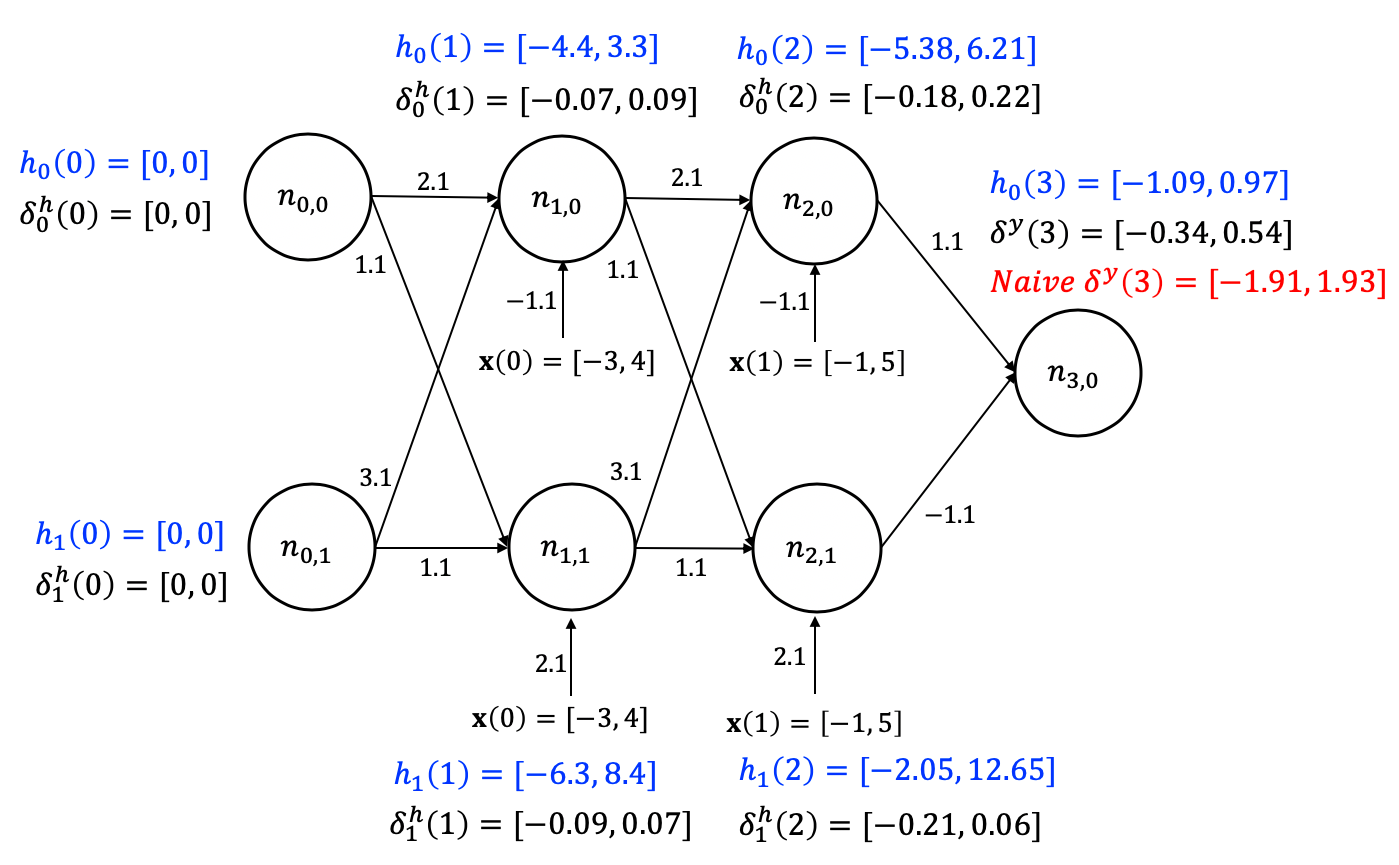}
\caption{ Interval analysis of a recurrent neural network with Sigmoid activation function.}
\label{fig:rnn_sigmoid}
\end{figure}

Fig.~\ref{fig:rnn_sigmoid} shows a \emph{many-to-1} Vanilla RNN, $f$, where all
neurons use the \Sigmoid{} activation. The entire RNN has 1
hidden layer of 2 neurons, receives a sequence of 2 inputs, and returns
a single output. For ease of presentation, the unrolled version of
this RNN is shown in Fig.~\ref{fig:rnn_sigmoid} for an input sequence of length 2.
Thus, $n_{t,i}$ denotes the $i^{th}$  node in the $t^{th}$ sequence (time step). The initial values in $\vHid(0)$ and $\diffh(0)$ are set to 0.
The goal is to bound $\diffy(3)$, the difference between outputs of the
original RNN $f$ and a modified RNN $f^\prime$; here,
$\diffy(3) = |f^\prime(\vInputSeq(0) ,\vInputSeq(1)) - f(\vInputSeq(0) ,\vInputSeq(1))|$.
In this example, the second RNN, $f^\prime$, is derived from the
original RNN $f$ by rounding its edge weights to the nearest whole
numbers.

The na\"{i}ve approach is to leverage an existing verification tool
such as \Popqorn{}~\cite{ko2019popqorn}, originally designed to
quantify the robustness of a single RNN.  As shown for the network in Fig.~
\ref{fig:rnn_diff_naive}, we can use \Popqorn{} to bound the output of
the combined network.  For our running example, the bounds computed by
\Popqorn{} are $\diffy(3) = [-1.91,1.93]$.  However, as our analysis
shows in this paper, the bounds are overly conservative.  The reason
is because, to soundly compute the difference $\diffy(3) =
[f^\prime_{low},f^\prime_{up}] - [f_{low},f_{up}] = [f^\prime_{low} -
  f_{up}, f^\prime_{up}-f_{low}]$, \Popqorn{} has to introduce
significant approximation error.

\DiffRNN{}, in contrast, overcomes this problem by pairing neurons and
edges of the first network with their counterparts of the second
network, and directly computing the difference intervals layer by
layer.  By directly computing the differences as opposed to the output bounds of the corresponding neurons, we hope to obtain much
tighter bounds.
However, there are unique challenges in directly bounding the differences. 
While bounding the non-linear activation function $y=\sigma(x)$ for
a single neuron is relatively easy~\cite{zhang2018efficient}, doing so
for a pair of neurons at the same time is more difficult because we
must bound $z= \sigma(x^\prime) -\sigma(x)$, which involves two variables.
%
While we could bound the individual terms $ \sigma(x^\prime) $ and 
$ \sigma(x) $, and then subtract their bounds, doing so introduces too
much approximation error.
%

To solve the problem, we propose the following new technique.  First, we rewrite
the difference as $z= \sigma(x+\delta_x) - \sigma(x)$, where $x$ is
the value of neuron's output in network $f$ and $\delta_x = (x^\prime-x)$ is the
difference between the outputs of two corresponding neurons in $f$ and $f'$. 
%
Given the intervals of $x$ and $\delta_x$, we then examine all
possible combinations of their upper and lower bounds, and match these
combinations with a set of pre-defined patterns, to soundly compute
the interval of $z$.

For LSTMs, directly bounding the difference $z=x^\prime \cdot
\sigmoid(x^\prime)-x \cdot \sigmoid(x)$ or $z=\tanh(x^\prime) \cdot
\sigmoid(x^\prime)-\tanh(x)\cdot \sigmoid(x)$ is even more
challenging.  To the best of our knowledge, no existing verification
tool for neural networks can compute tight linear bounds for such
functions.
Our solution is to formulate them as constrained optimization problems
and solve these problems using global optimization
tools~\cite{price2006differential}.  In addition, we prove the
soundness of these bounds using \emph{dReal}, an off-the-shelf \emph{delta-sat} SMT solver
that supports nonlinear constraints.

For the running example, our method would be able to compute the
bounds $\diffy(3) = [-0.34,0.54]$, which is more than 3X tighter than
the bounds computed by \Popqorn{}.
The complete results of our experimental comparison with \Popqorn{} will be presented in Section~\ref{sec:experiments}.

\section{Directly Computing the Difference Interval}
Our method for verifying Vanilla RNNs is
shown in Algorithm~\ref{alg:d_v_rnn}. It takes
two networks $f$ and $f^\prime$, the input region $X$, and a small
$\epsilon$ as input. 
After initializing the hidden state and the
difference interval, it computes $\vHid(t)$ and
$\diffh(t)$ of the subsequent layers by applying the affine
transformation (i.e., multiplying by the edge weights) followed by performing 
the non-linear transformation, whose details will be presented in Algo.~\ref{alg:non-linear}.  This is repeated layber by layer, 
until the output layer is reached.  In the end, it computes the final difference interval from $\diffh(T)$ and $\vHid(T)$.  
As mentioned earlier, we leverage the existing tool \Popqorn{}~\cite{ko2019popqorn} to compute the intervals $\vHid(t)$, while focusing on computing tight bounds on  the differences $\diffa(t)$ and $\diffh(t)$. 
%

{\small
\begin{algorithm}[t]
\small
\caption{Differential Verification of Vanilla RNNs. \label{alg:d_v_rnn}}
\SetKwProg{Fn}{Function}{:}{}

\KwIn{\textit{network f, network $f^\prime$, input region X}}
\KwOut{$\diffy(T)$}
\vspace{0.2cm}
\textbf{Init:} Initialize $[L(\vHid(0)), U(\vHid(0))]$ and $[\underline{\diffh}(0), \overline{\diffh}(0)]$ to 0\;
\For{$t$ : 1 to $T$}{
{\color{teal} \tcp{affine transformer}}
Compute $[L(\vAff(t)), U(\vAff(t))]$ and $[\underline{\diffa}(t), \overline{\diffa}(t)]$;\\
{\color{teal} \tcp{nonlinear transformer (Algo.~\ref{alg:non-linear})}}
Compute $[L(\vHid(t)), U(\vHid(t))]$ and $[\underline{\diffh}(t), \overline{\diffh}(t)]$;\\

}
Compute $[\underline{\diffy}(T), \overline{\diffy}(T)]$; 
\end{algorithm}}

\subsection{Affine Transformer}

For Vanilla RNNs, the affine transformation computes each $\diffa(t)$
in two parts. 
The first part is caused by the differences between the edge weights
in $\whx$, denoted $\whx^{\Delta}$, for edges connecting the
current input to neurons:

{\small
\begin{flalign*}
\bm{\delta_t}(\whx) &= \whx' \cdot \vInputSeq(t) - \whx \cdot\vInputSeq (t) \\&=\whx^{\Delta} \cdot \vInputSeq(t)
\end{flalign*}
}

The second part is caused by the differences between the edge weigths
in $\whh$, denoted $\whh^{\Delta}$, for edges connecting the previous
hidden states to current hidden states, as well as the differences
included in the previous hidden states, denoted $\diffh(t-1)$.

{\small
\begin{flalign*}
\bm{\delta_t}(\whh) &= \whh' \cdot \vHid'(t-1) - \whh \cdot \vHid(t-1)  \\
\end{flalign*}
}

Adding $\whh' \cdot \vHid(t-1)$ to the first term and subtracting it from the second term, we get:

{\small
\begin{flalign*}
\bm{\delta_t}(\whh) &= \whh' \cdot \vHid'(t-1) - \whh \cdot \vHid(t-1) \\
&+({\color{red} \whh' \cdot \vHid(t-1) -\whh' \cdot \vHid(t-1) })\\
&=(\whh' \cdot \vHid'(t-1)-{\color{red} \whh' \cdot \vHid(t-1)})\\
&+({\color{red} \whh' \cdot \vHid(t-1)}- \whh \cdot \vHid(t-1))\\&=
\whh' \cdot \diffh(t-1)+\whh^{\Delta} \cdot \vHid(t-1)
\end{flalign*}
}

$\diffa(t)$ is then the sum of these two parts:

{\small
\begin{flalign*}
\diffa(t) &= \bm{\delta_t}(\whx)  + \bm{\delta_t}(\whh)
\end{flalign*}
}

For LSTMs, the high-level verification procedure is similar to
Algo.~\ref{alg:d_v_rnn} and is formalized in
Algo.~\ref{alg:d_lstm} in Appendix. The differences for gate $\bm{v}
\in \{\igate, \fgate, \ggate, \ogate\} $ within each LSTM cell is
computed as follows: $\diffv(t) = \bm{\delta_t}(\wvx)  +
\bm{\delta_t}(\wvh)$.

\ignore{
{\small
\begin{flalign*}
\diffi(t) &= \bm{\delta_t}(\wix)  + \bm{\delta_t}(\wih)\\
\difff(t) &= \bm{\delta_t}(\wfx)  + \bm{\delta_t}(\wfh)\\
\diffg(t) &= \bm{\delta_t}(\wgx)  + \bm{\delta_t}(\wgh)\\
\diffo(t) &= \bm{\delta_t}(\wox)  + \bm{\delta_t}(\woh)
\end{flalign*}
}
}

\subsection{Nonlinear Transformer}
\mypara{Vanilla RNN} 
Here, we define the activation function transformations to compute $\diffh(t)$ from $\diffa(t)$. We do so by rewriting the following equation, using the definition of $\diffh(t)$:

{\small
\begin{flalign*}
\diffh(t) &= \vHid'(t)- \vHid(t)\\&=
\sigma(\vAff'(t))- \sigma(\vAff(t)) \\&=
\sigma(\vAff(t)+\diffa(t) ) -  \sigma(\vAff(t))
\end{flalign*}
}

where $\sigma$ is the nonlinear activation function. While \ReluDiff{}~\cite{paulsen2020reludiff} solves this problem for $\sigma = ReLU$, by  exploiting the piece-wise linearity of $ReLU$, we propose new techniques for $\sigma = Sigmoid$ or $Tanh$, as well as composite nonlinear operations built upon them. Note that the technique can be used for other types of monotonic functions as well.

To obtain the tightest linear bounds on $\diffh(t)$, we formulate this problem as two optimization problems:

{\small
\begin{equation*}
\begin{aligned}
\underline{\diffh}(t)=
& \underset{\vAff(t),\diffa(t)}{\text{minimize}}
& & \sigmoid(\vAff(t)+\diffa(t) ) -  \sigmoid(\vAff(t)) \\
& \text{subject to}
& & \vAff(t) \in [\underline{\vAff}(t),\overline{\vAff}(t)], \;\diffa(t) \in [\underline{\diffa}(t),\overline{\diffa}(t)]
\end{aligned}
\end{equation*}}

{\small
\begin{equation*}
\begin{aligned}
\overline{\diffh}(t)=
& \underset{\vAff(t),\diffa(t)}{\text{maximize}}
& & \sigmoid(\vAff(t)+\diffa(t) ) -  \sigmoid(\vAff(t)) \\
& \text{subject to}
& & \vAff(t) \in [\underline{\vAff}(t),\overline{\vAff}(t)], \;\diffa(t) \in [\underline{\diffa}(t),\overline{\diffa}(t)]
\end{aligned}
\end{equation*}}

%
These are two-variable optimization problems of the form $f(x,d)
= \sigmoid(x+d)-\sigmoid(x)$, which are expensive to solve at run
time.  To reduce the computational cost, we propose to reduce them
first to single-variable optimization problems, by leveraging the fact
that $ f(x,d) $ is monotonic with respect to $ d $.
Our goal is to compute the maximum and minimum of $f(x,d)
= \sigmoid(x+d)-\sigmoid(x)$, where $x\in[x_l,x_u]$ and
$d\in[d_l,d_u]$.  Due to the monotonicity of $ f(x,d) $ with respect
to $d$, we know that the minimum always occurs when $d = d_l$ and the
maximum occurs when $d = d_u$. Thus, the problem is reduced to finding
the maximum and minimum of $f(x,d)$ for a fixed $d=d_l$ or $d=d_u$.

\begin{figure}[ht]
  \centering
  \begin{subfigure}[b]{0.51\linewidth}
    \centering\includegraphics[width=130pt]{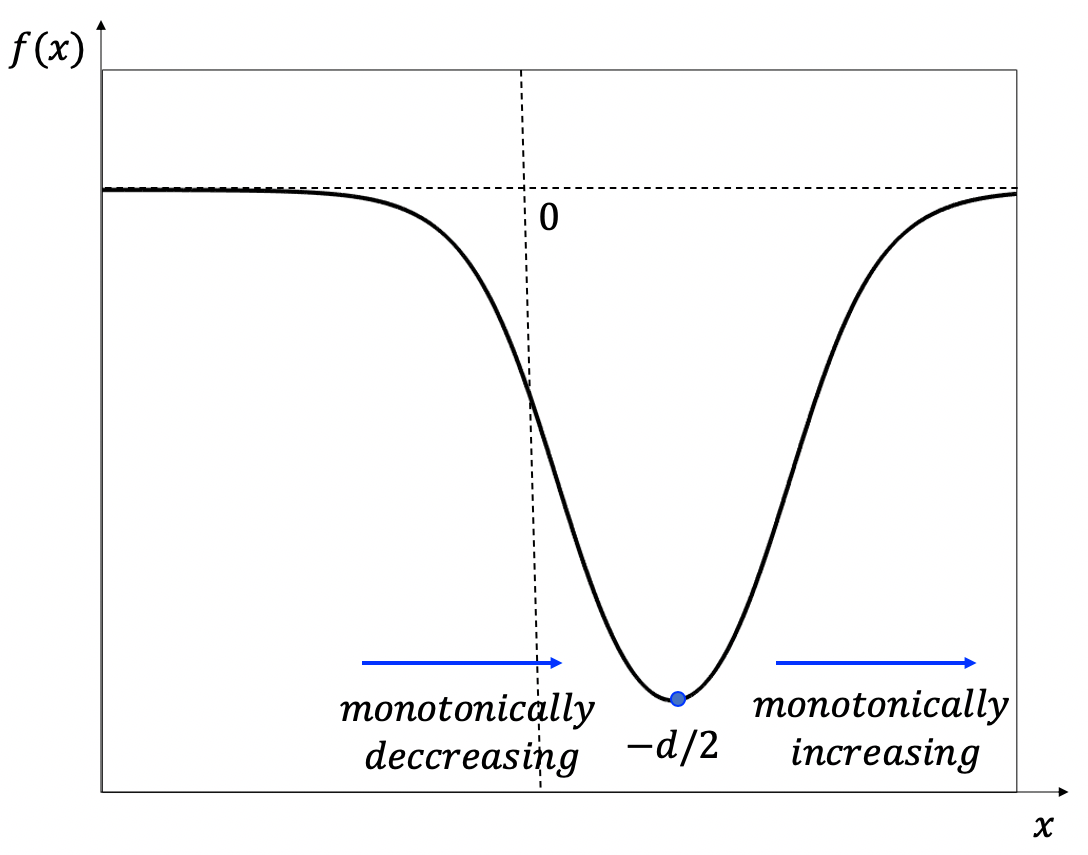}
    \caption{\label{fig:sigmoid_case1} $d < 0$}
  \end{subfigure}%
  \begin{subfigure}[b]{0.5\linewidth}
    \centering\includegraphics[width=130pt]{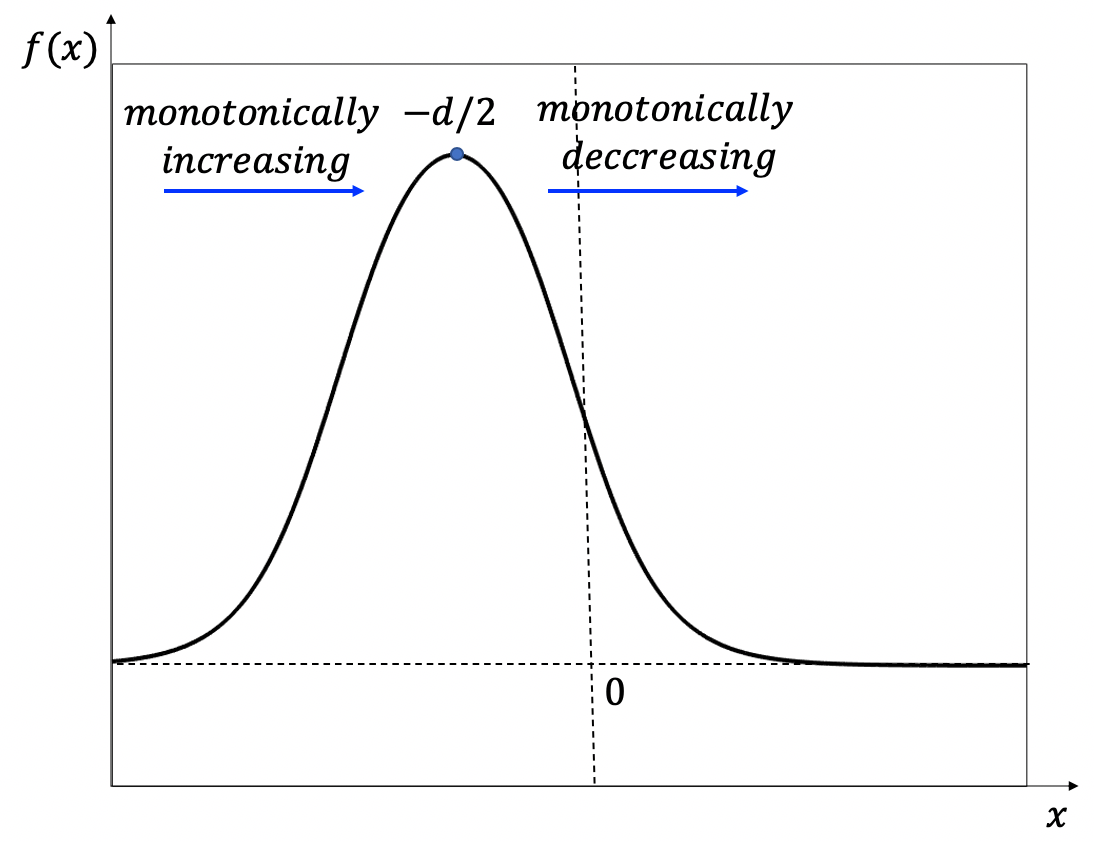}
    \caption{\label{fig:sigmoid_case3} $d\geq0$}
  \end{subfigure}
  \caption{$f(x) = \sigmoid(x+d)-\sigmoid(x)$. (\subref{fig:sigmoid_case1}) shows $f(x)$ for $d < 0$ and (\subref{fig:sigmoid_case3}) shows  $f(x)$ for $d \geq 0$.}
 \label{fig:sigmoid}
\end{figure}

Depending on the actual value of $d$ being either positive or
negative, the function $f(x,d)$ will be one of the two forms
illustrated in Fig.~\ref{fig:sigmoid}.

Thus, to compute the minimum of $f(x,d)$, there are three cases to
consider when $ d_l $ is positive, and another three cases to consider
when $ d_l $ is negative. 
For instance, when $d_l \leq 0$ (Fig.~\ref{fig:sigmoid_case1}), if
$x_l \leq x_u \leq -d_l/2$, since $f(x,d_l)$ is monotonically
decreasing in this region, we have $\min(f(x,d_l))=f(x_u,d_l)$; if
$x_l \leq -d_l/2\leq x_u$, we have $\min(f(x,d_l))=f(-d_l/2,d_l)$; and
if $-d_l/2\leq x_l \leq x_u$ (monotonically increasing), we have
$\min(f(x,d_l))=f(x_l,d_l)$. 
The main advantage is that, by plugging in the bounds of $x$ and $d$,
we get the bounds of $f(x,d)$ immediately, without the need to solve
any optimization problem at run time.

To compute $\underline{\diffh}(t)$ and $\overline{\diffh}(t)$, there
are 12 cases in total, the details of which are formalized in
Algo.~\ref{alg:non-linear}. For $Tanh$ ($\tanh$), the 12 cases are
exactly the same as those for $Sigmoid$ ($\sigmoid$); thus, we omit
them for brevity\footnote{For non-monotonic activation functions we can compute the maximum and minimum using off-the-shelf global optimization tools and then validate the computed bounds using SMT solvers.}.

\begin{algorithm}[t]
\small
\caption{Over-approximating Non-linear Activation.\label{alg:non-linear}}
\SetKwProg{Fn}{Function}{:}{}

\KwIn{ value $[L(\vAff(t)), U(\vAff(t))]$, difference $[\underline{\diffa}(t),\overline{\diffa}(t)]$}
\KwOut{value $[L(\vHid(t)), U(\vHid(t))]$, difference $[\underline{\diffh}(t),\overline{\diffh}(t)]$}
\vspace{0.2cm}
{\color{teal}\tcp{Computing $\vHid(t)$}}
{\color{teal}\tcp{Using a previous work \Popqorn{}}}
$[L(\vHid(t)), U(\vHid(t))]$ =  \Popqorn{}.NonLinearTransformer($[L(\vAff(t)), U(\vAff(t))]$);
{\color{teal}\tcp{Computing $\underline{\diffh}(t)$}}

\If{$\underline{\diffa}(t)\geq 0$}{
\vspace{0.2cm}
\scalebox{0.85}{
\begin{minipage}{\linewidth}

\uIf{$\overline{\vAff}(t) \leq -\underline{\diffa}(t)/2$}{
\vspace{0.2cm}
    $\underline{\diffh}(t) = \sigmoid(\underline{\vAff}(t)+ \underline{\diffa}(t))-\sigmoid(\underline{\vAff}(t))$;
    \vspace{0.2cm}
  }
  \uElseIf{$\underline{\vAff}(t) \geq -\underline{\diffa}(t)/2$}{
  \vspace{0.2cm}
        $\underline{\diffh}(t) = \sigmoid(\overline{\vAff}(t)+ \underline{\diffa}(t))-\sigmoid(\overline{\vAff}(t))$;
  }
  \Else{
    $a =\sigmoid(\overline{\vAff}(t)+ \underline{\diffa}(t))-\sigmoid(\overline{\vAff}(t))$;
  
 $b = \sigmoid(\underline{\vAff}(t)+ \underline{\diffa}(t))-\sigmoid(\underline{\vAff}(t))$;

    $\underline{\diffh}(t) = \min(a,b)$;
  }
\end{minipage}}

}
\Else{

\vspace{0.2cm}
\scalebox{0.85}{
\begin{minipage}{\linewidth}

\uIf{$\overline{\vAff}(t) \leq -\underline{\diffa}(t)/2$}{
\vspace{0.2cm}
    $\underline{\diffh}(t) = \sigmoid(\overline{\vAff}(t)+ \underline{\diffa}(t))-\sigmoid(\overline{\vAff}(t))$;
    \vspace{0.2cm}
  }
  \uElseIf{$\underline{\vAff}(t) \geq -\underline{\diffa}(t)/2$}{
  \vspace{0.2cm}
        $\underline{\diffh}(t) = \sigmoid(\underline{\vAff}(t)+ \underline{\diffa}(t))-\sigmoid(\underline{\vAff}(t))$;
  }
  \Else{
  
          $\underline{\diffh}(t) =\sigmoid(-\underline{\diffa}(t)/2+ \underline{\diffa}(t))-\sigmoid(-\underline{\diffa}(t)/2)$;
 
  }
\end{minipage}}

}
{\color{teal}\tcp{Computing $\overline{\diffh}(t)$}}
\If{$\overline{\diffa}(t)\geq 0$}{
\vspace{0.2cm}
\scalebox{0.85}{
\begin{minipage}{\linewidth}

\uIf{$\overline{\vAff}(t) \leq -\overline{\diffa}(t)/2$}{
\vspace{0.2cm}
    $\overline{\diffh}(t) = \sigmoid(\overline{\vAff}(t)+ \overline{\diffa}(t))-\sigmoid(\overline{\vAff}(t))$;
    \vspace{0.2cm}
  }
  \uElseIf{$\underline{\vAff}(t) \geq -\overline{\diffa}(t)/2$}{
  \vspace{0.2cm}
        $\overline{\diffh}(t) =\sigmoid(\underline{\vAff}(t)+ \overline{\diffa}(t))-\sigmoid(\underline{\vAff}(t))$;
  }
  \Else{
  
            $\overline{\diffh}(t) = \sigmoid(-\overline{\diffa}(t)/2+ \overline{\diffa}(t))-\sigmoid(-\overline{\diffa}(t)/2)$;
   
  }
\end{minipage}}

}
\Else{

\vspace{0.2cm}
\scalebox{0.85}{
\begin{minipage}{\linewidth}

\uIf{$\overline{\vAff}(t) \leq -\overline{\diffa}(t)/2$}{
\vspace{0.2cm}
    $\overline{\diffh}(t) = \sigmoid(\underline{\vAff}(t)+ \overline{\diffa}(t))-\sigmoid(\underline{\vAff}(t))$;
    \vspace{0.2cm}
  }
  \uElseIf{$\underline{\vAff}(t) \geq -\overline{\diffa}(t)/2$}{
  \vspace{0.2cm}
        $\overline{\diffh}(t) = \sigmoid(\overline{\vAff}(t)+ \overline{\diffa}(t))-\sigmoid(\overline{\vAff}(t))$;
  }
  \Else{
  
      $a =\sigmoid(\overline{\vAff}(t)+ \overline{\diffa}(t))-\sigmoid(\overline{\vAff}(t))$;
  
 $b =\sigmoid(\underline{\vAff}(t)+ \overline{\diffa}(t))-\sigmoid(\underline{\vAff}(t))$;

    $\overline{\diffh}(t) = \max(a,b)$;
   
  }
\end{minipage}}

}
\end{algorithm}


The final difference interval for Vanilla RNNs is computed from $\diffh(T)$ and $\vHid(T)$ as follows:

{\small
$$\diffy =  \whout' \cdot \diffh (T) + \whout^\Delta\cdot \vHid(T)$$
}

\mypara{LSTM}
Next, we consider computing $\diffh(t)$ for LSTMs. First, we need to compute  $\diffc(t)$. In the following computations, we add underscore ``$in$'' to denote the value of each variable after the affine transform but before the nonlinear activation. Based on the definition of  $\diffc(t)$, we have

{\small
\begin{flalign}
\diffc(t)= \cstate'(t)-\cstate(t)
\label{deltac}
\end{flalign}
}

,where $\cstate'(t)$ and $\cstate(t)$ are intervals of the cell states for $t^{th}$ sequence. Based on the definition of the cell state, we have 

{\small
\begin{flalign*}
\cstate'(t) &= \sigmoid(\fgate'_{\uin}(t)) \odot \cstate'(t-1) + \sigmoid(\igate'_{\uin}(t)) \odot  \tanh(\ggate'_{\uin}(t))\\
\cstate(t) &= \sigmoid(\fgate_{\uin}(t)) \odot \cstate(t-1) + \sigmoid(\igate_{\uin}(t)) \odot  \tanh(\ggate_{\uin}(t))
\end{flalign*}
}

After replacing the above equation for $\cstate'(t)$ and $\cstate(t)$ in (\ref{deltac}) and simplification, we have 

{\small
\begin{flalign*}
\diffc(t)&= \sigmoid(\fgate'_{\uin}(t)) \odot \cstate'(t-1) -\sigmoid(\fgate_{\uin}(t)) \odot \cstate(t-1) \\
&+\sigmoid(\igate'_{\uin}(t)) \odot  \tanh(\ggate'_{\uin}(t))-
\sigmoid(\igate_{\uin}(t)) \odot  \tanh(\ggate_{\uin}(t)),
\end{flalign*}
}

where each variable is an interval. Since the above equation contains
eight variables, manually enumerating all possible solutions at the
design time is practically infeasible. Thus, we have to compute them
at run time. To make the problem tractable, we divide $\diffc(t)$
into two parts, each with four variables, and optimize them
independently.  We define the two parts as follows: 

{\small
\begin{flalign*}
\diffc_1(t)&= \sigmoid(\fgate'_{\uin}(t)) \odot \cstate'(t-1) -\sigmoid(\fgate_{\uin}(t)) \odot \cstate(t-1)\\
\diffc_2(t)&= \sigmoid(\igate'_{\uin}(t)) \odot  \tanh(\ggate'_{\uin}(t))-
\sigmoid(\igate_{\uin}(t)) \odot  \tanh(\ggate_{\uin}(t))
\end{flalign*}}

Since $\fgate'_{\uin}(t) = \fgate_{\uin}(t) + \difff_{\uin}(t)$,
$\igate'_{\uin}(t) = \igate_{\uin}(t) + \diffi_{\uin}(t)$,
$\ggate'_{\uin}(t) = \ggate_{\uin}(t) + \diffg_{\uin}(t)$ and
$\cstate'(t-1) = \cstate(t-1)+ \diffc_{\uin}(t-1)$, we can rewrite
$\diffc_1(t)$ and $\diffc_2(t)$ as follows:

{\small
\begin{eqnarray*}
\diffc_1(t)&=& \sigmoid(\fgate_{\uin}(t) + \difff_{\uin}(t))) \odot (\cstate(t-1)+ \diffc_{\uin}(t-1)) \\
           & &-\sigmoid(\fgate_{\uin}(t)) \odot \cstate(t-1) \\
\diffc_2(t)&=& \sigmoid(\igate_{\uin}(t) + \diffi_{\uin}(t)) \odot  \tanh(\ggate_{\uin}(t) + \diffg_{\uin}(t))\\
           &  &-\sigmoid(\igate_{\uin}(t)) \odot \tanh(\ggate_{\uin}(t)) \\
\overline{\diffc}(t) & = & \overline{\diffc_1}(t)+\overline{\diffc_2}(t)\\
\underline{\diffc}(t) & = &\underline{\diffc_1}(t)+\underline{\diffc_2}(t) \\
\end{eqnarray*}
}

To reduce notational complexity, we rewrite the above equations as
$f_1(x,d_x,y,d_y)=  \sigmoid (x+d_x) \cdot (y+d_y) -  \sigmoid(x)
\cdot y$ and $f_2(x,d_x,y,d_y)=  \sigmoid (x+d_x) \cdot \tanh(y+d_y) -
\sigmoid(x) \cdot \tanh(y)$. Observing that both equations are
monotonic with respect to $ d_y $, we convert the four-variable
optimization problems to three variables. We first rewrite $ f_1 $ as:

{\small
\begin{flalign*}
f_1(x,d_x,y,d_y) &= \sigmoid(x+d_x) \cdot y + \sigmoid (x+d_x) \cdot d_y -  \sigmoid(x) \cdot y \\&= (\sigmoid (x+d_x) - \sigmoid(x)) \cdot y + \sigmoid (x+d_x) \cdot d_y 
\end{flalign*}}

Now, since $\sigmoid (x+d_x) \geq 0$, the maximum of $f_1$ occurs when
$d_y = \overline{d_y}$ and the minimum of $f_1$ happens at $d_y =
\underline{d_y}$, we can treat $ d_y $ as a constant. Similarly, for
$f_2$, since $\sigmoid (x+d_x)  \geq 0$ and $ \tanh(y+d_y)$ is
monotonically increasing, the maximum and minimum of $f_2$ happens
when $d = \overline{d_y}$ and $d = \underline{d_y}$, respectively.
Finally, to obtain $\diffh(t)$, we solve two 4-variable optimization
problems:

{\small
\begin{eqnarray*}
\diffh(t) &=&
 \sigmoid(\ogate_{\uin}(t)+\diffo_{\uin}(t))  \odot \tanh(\cstate(t)+\diffc (t))\\
          & &- \sigmoid(\ogate_{\uin}(t))  \odot \tanh(\cstate(t)) \\
\overline{\diffh}(t) &=& \max(\diffh(t))\\
\underline{\diffh}(t) &=& \min(\diffh(t))
\end{eqnarray*}
}

These functions are again of the form $f_2(x,d_x,y,d_y)= \sigmoid
(x+d_x) \cdot \tanh(y+d_y) - \sigmoid(x) \cdot \tanh(y)$ and thus can
be converted to 3-variable optimization problems.  We use
off-the-shelf global optimization tools, which are written in Python
and based on differential evolution (DE) \cite{price2006differential}, to solve these optimization
problems. In evolutionary computation, DE is a method that optimizes a problem by iteratively trying to improve a candidate solution with regard to a given measure of quality.
After that, we use \emph{dReal} \cite{gao2013dreal}, which is a \emph{delta-sat} SMT
solver with support for nonlinear functions, to validate the computed
bounds. If the bounds are not yet sound according to \emph{dReal}, we
slightly increase the maximum or decrease the minimum until they are
proved to be sound (see Algo.~\ref{alg:non-linear_lstm} in Appendix).
The final difference interval for LSTMs is derived in a way that is
similar to Vanilla RNNs. The details are formalized in  Algo.~\ref{alg:d_lstm} and Algo.~\ref{alg:non-linear_lstm} in Appendix.

\section{Experiments}
\mypara{Benchmarks}
Our benchmarks are 12 feed forward neural networks with $Sigmoid$ and $Tanh$ activations \footnote{One of our evaluation objectives  is to extend the results of \ReluDiff{} to general activation functions instead of \emph{ReLUs}. This also allows us to validate our methodology in the relatively simpler world of feedforward networks before tackling RNNs.}, 12 Vanilla RNNs, and 6 LSTMs trained using the MNIST \cite{lecun-mnisthandwrittendigit-2010} and Human Activity Recognition (HAR) \cite{anguita2013public} data sets. 
From each network $f$, we produce $f^\prime$ by rounding the edge weights of $f$ from 32-bit floats to 16-bit floats. We generate the input regions for differential verification using global perturbation~\cite{SinghGPV19} or targeted pixel perturbation~\cite{GopinathKPB18}. We randomly take 100 test inputs, and for each one, we allow each of the inputs to be perturbed $-/+1\%$ of the whole range (\emph{global} perturbation), or we randomly pick \emph{3 inputs} and set their range to the whole range (targeted perturbation). Given an input region, the goal is to verify the difference of at most $\epsilon$ between the outputs of $f$ and $f^\prime$. The value of $\epsilon$ is specified for each benchmark separately.

\myipara{MNIST}
MNIST is one of the most popular data sets for handwritten digits recognition, consisting of 60,000 and 10,000 images corresponding to training and test data. The images are 28x28 = 784 pixels, and each pixel has a grayscale value in the range [0, 255] which is usually scaled to [-1,   1]. The neural networks trained on this data set generate 10 outputs typically in the range [-10, 10] and the digit with the highest score
is the chosen classifcation. 

\myipara{Human Activity recognition}
HAR is a labeled time-series data set used to train models for
human activity recognition. The data is recorded from accelerometer and gyroscope sensors in waist-mounted smartphones.  In total, 561 input statistics are computed from these two sensors including max, min, mean, etc.,  which are normalized to the range [-1, 1]. This data is obtained from the recordings of 30 subjects performing six activities:  walking, walking upstairs, walking downstairs, sitting, standing, and laying down. The network trained on this data set takes 561 inputs and generates 6 outputs typically in the range [-20, 20].  The output with the maximum value is the predicted class.

\mypara{Experimental Evaluation}
We run the experiments on an Intel Core-i7 Macbook Pro with 2.7 GHz processors and 16 GB RAM.  Timeout for each verifcation problem is set
to 30 minutes. We compare the results of feed-forward networks with {\scshape Crown} \cite{zhang2018efficient} which is the state-of-the-art verification tool for a single feed-forward neural network. 

Among the existing tools for verifying a single RNN~\cite{
  jia2019certified, shi2020robustness}, we find empirically that
{\scshape Popqorn} is significantly more accurate than those
from~\cite{ jia2019certified, shi2020robustness}: the bounds computed
by \cite{ jia2019certified, shi2020robustness} often have too much approximation error, and hence would give too many false positives for the differential verification problem. Therefore, we compare our experimental results on RNNs
with {\scshape Popqorn}, which leverages gradient descent techniques
to compute linear bounds on nonlinear surfaces $x \cdot \sigmoid(y)$
and $\sigmoid(x) \cdot \tanh(y)$.  As {\scshape Popqorn} evaluates bounds using numerical tools based on gradient descent, while the approach is sound in theory, it is susceptible to numerical precision issues. Hence, we added an extra validation
step using \emph{dReal} to ensure numerical precision of the bounds computed
by {\scshape Popqorn}.

\mypara{Results}
In the 3000 differential verification problems that we consider, {\scshape DiffRNN}  can verify 2887 out of 3000 problems and is faster than {\scshape Crown} and {\scshape Popqorn} in more than $93\%$ of the cases. {\scshape Crown} and {\scshape Popqorn} in total can verify only 1140 out of 3000 problems. {\scshape DiffRNN} returns \emph{Unknown} for other 123 verification problems that cannot verify.

Table.~\ref{tab:dnn_mnist} shows the results of differential
verification of feed-forward neural networks with $Sigmoid$ activation
trained on MNIST data set. The networks have 3 structures $3\times128$
(3 hidden layers of 128 neurons), $2\times512$ (2 hidden layers of 512
neurons) and $4\times1024$ (4 hidden layers of 1024 neurons). Thus,
the networks have 2, 3 and 4 layers in addition to input and output
layers. The goal is to verify the difference of at most 1 ($\epsilon =
1$) between the outputs of $f$ and $f^\prime$.  Among the 600
verification problems shown in Table.~\ref{tab:dnn_mnist}, {\scshape
DiffRNN} verified all of them while {\scshape Crown} verified only
224.  

Table.~\ref{tab:dnn_har} shows the results of differential
verification of feed-forward neural networks with $Tanh$ activation
and three types of structures: $3\times128$, $2\times1024$ and
$4\times512$ on the HAR data set.  {\scshape DiffRNN} verified 591 of
the 600 cases for $\epsilon = 2$ as apposed to the 282 cases verified
by {\scshape Crown}.  

Table.~\ref{tab:rnn_mnist} shows the results of verifying Vanilla RNNs
with $\epsilon=1$ on the MNIST data set. The network structures are $4
\times 128$ (4 sequences of 128 neurons), $7 \times 32$ (7 sequences of 32
neurons) and $14 \times 8$ (14 sequences of 8 neurons). Among the 600
verification problems, {\scshape DiffRNN} verified 502 while {\scshape
  Popqorn}  only verified 156. 

Table.~\ref{tab:rnn_har} shows the results of verifying Vanilla RNNs
on the HAR data set.  The networks are $3 \times 32$, $3 \times 128$
and $11 \times 8$ and $\epsilon=2$.  {\scshape DiffRNN} was faster
than {\scshape Popqorn} in all cases and also verified more properties
(598/600), while {\scshape Popqorn} only verified 478 properties. 

Finally, Table.~\ref{tab:lstm_har} shows the results of verifying
LSTMs trained on the HAR data set for structures $3 \times 32$, $3
\times 64$ and $11 \times 8$ and $\epsilon=0.1$. The results again
show that {\scshape DiffRNN} was better: it verified all 600 cases
while {\scshape Popqorn} can verify none of them.

\begin{table}[ht]
\centering
\begin{tabular*}{.5\textwidth}{@{\extracolsep{\fill}}llll}
\toprule
Benchmark & DIFFRNN (New) & CROWN & Avg. Speedup\\
\midrule
dnn-3x128-global&100/100, 22.7s&1/100, 25.9s&1.14\\
dnn-2x512-global&100/100, 52.4s&7/100, 61.9s&1.18\\
dnn-4x1024-global&100/100, 207.8s&0/100, 246.9s&1.27\\
dnn-3x128-3-inputs&100/100, 16.5s&79/100, 18.9&1.14\\
dnn-2x512-3-inputs&100/100, 44.5s&100/100, 49.8s&1.11\\
dnn-4x1024-3-inputs&100/100, 197.6s&37/100, 204.9s&1.03\\[1ex]
\bottomrule
\end{tabular*}
\caption{Verified problems, totall verification problems and Avg verification time (seconds) $(\cdot /\cdot),\cdot$ of DIFFRNN and CROWN on MNIST for DNN with Sigmoid activation and $\epsilon = 1$.
 \label{tab:dnn_mnist}}
\end{table}

\begin{table}[ht]
\centering
\begin{tabular*}{.5\textwidth}{@{\extracolsep{\fill}}llll}
\toprule
Benchmark & DIFFRNN(New) & CROWN & Avg. Speedup\\
\midrule
dnn-3x128-global&100/100, 18.4s&51/100, 22.2s&1.20\\
dnn-2x1024-global&100/100,118.5s&33/100, 147.4s&1.24\\
dnn-4x512-global&91/100, 160.2s&0/100, 218.2s&1.36\\
dnn-3x128-3-inputs&100/100, 16.0s&98/100, 19.0s&1.19\\
dnn-2x1024-3-inputs&100/100, 96.1s&100/100, 122.1s&1.27\\
dnn-4x512-3-inputs&100/100, 123.2s&0/100, 162.3s&1.31\\[1ex]
\bottomrule
\end{tabular*}
\caption{Verified problems, totall verification problems and Avg verification time (seconds) $(\cdot /\cdot),\cdot$ of DIFFRNN and CROWN on HAR for DNN with Tanh activation and $\epsilon = 2$.
 \label{tab:dnn_har}}
\end{table}

\begin{table}[ht]
\centering
\begin{tabular*}{.5\textwidth}{@{\extracolsep{\fill}}llll}
\toprule
Benchmark & DIFFRNN (New) & POPQORN & Avg. Speedup\\
\midrule
rnn-4x128-global&100/100, 562.9s&0/100, 1148.7s&2.04\\
rnn-7x32-global&57/100, 75.1s&0/100,136.1s&1.81\\
rnn-14x8-global&50/100, 14.1s&15/100, 24.4s&1.72\\
rnn-4x128-3-inputs&100/100, 571.8s&54/100, 1121.9s&1.96\\
rnn-7x32-3-inputs&100/100, 74.2s&15/100, 128.3s&1.72\\
rnn-14x8-3-inputs&95/100, 16.2s&57/100, 24.3s&1.49\\[1ex]
\bottomrule
\end{tabular*}
\caption{Verified problems, totall verification problems and Avg verification time (seconds) $(\cdot /\cdot),\cdot$ of DIFFRNN and POPQORN on MNIST for Vanilla RNN with Tanh activation and $\epsilon = 1$.
 \label{tab:rnn_mnist}}
\end{table}

\begin{table}[ht]
\centering
\begin{tabular*}{.5\textwidth}{@{\extracolsep{\fill}}llll}
\toprule
Benchmark & DIFFRNN (New) & POPQORN & Avg. Speedup\\
\midrule
rnn-3x32-global&100/100, 28.4s&60/100, 58.5s&2.05\\
rnn-3x128-global&100/100, 424.8s&74/100, 849.0s&1.99\\
rnn-11x8-global&100/100, 11.6s&62/100, 18.0s&1.54\\
rnn-3x32-3-inputs&100/100, 29.0s&99/100, 57.8s&1.98\\
rnn-3x128-3-inputs&100/100, 424.8s&98/100, 825.5s&1.94\\
rnn-11x8-3-inputs&98/100, 11.6s&85/100, 17.3s&1.48\\[1ex]
\bottomrule
\end{tabular*}
\caption{Verified problems, totall verification problems and Avg verification time (seconds) $(\cdot /\cdot),\cdot$ of DIFFRNN and POPQORN on HAR for Vanilla RNN with Tanh activation and $\epsilon = 2$.
 \label{tab:rnn_har}}
\end{table}

\begin{table}[ht]
\centering
\begin{tabular*}{.5\textwidth}{@{\extracolsep{\fill}}llll}
\toprule
Benchmark & DIFFRNN (New) & POPQORN & Avg. Speedup\\
\midrule
lstm-3x32-global&100/100, 15512.6s&0/100, 16020.2s&1.03\\
lstm-3x64-global&100/100, 23602.9s&0/100, 18828.5s&0.79\\
lstm-11x8-global&100/100, 34766.3s&0/100, 35519.1s&1.02\\
lstm-3x32-3-inputs&100/100, 11820.0s&0/100, 14716.0s&1.24\\
lstm-3x64-3-inputs&100/100, 18789.0s&0/100, 17868.7s&0.95\\
lstm-11x8-3-inputs&100/100, 23224.1s&0/100, 33398.0s&1.43\\[1ex]
\bottomrule
\end{tabular*}
\caption{Verified problems, totall verification problems and Avg verification time (seconds) $(\cdot /\cdot),\cdot$ of DIFFRN and POPQORN on HAR for LSTM and $\epsilon = 0.1$.
 \label{tab:lstm_har}}
\end{table}

We also applied differential verification to 3 LSTM structures $4
\times 128$, $7 \times 32$ and $14 \times 8$ on the MNIST data
set. Neither \DiffRNN{} nor \Popqorn{} could verify any of these
problems. The reason behind the failure of \DiffRNN{} is that, after 2
sequences of propagation, the difference intervals start to get loose, 
resulting in poor performance of the \emph{dReal} SMT solver. 
As \emph{dReal} is an interval constraint based solver, 
bigger intervals requires \emph{dReal} to reason over bigger regions of space.
While \Popqorn{} was actually faster than \DiffRNN{} in this
experiment, the final differences it computed are too loose and cannot verify
the equivalence of any two networks for $\epsilon \leq 20$.  
Thus, the results point to directions for future research.

While in general \DiffRNN{} achieves a significant performance gain
compared to the state-of-the-art verification tools for a single RNN,
such as {\scshape Popqorn}, it can only tightly bound similarity of
two 1-layer RNNs with up to 20 input sequences. To the best of our
knowledge, there is no existing technique based on IBP that can
certify multi-layer RNNs with long input sequences. {\scshape
  Popqorn}, \cite{jia2019certified} and \cite{shi2020robustness} that
are tools for quantifying the robustness of a single RNN can only
certify the robustness for 1-layer RNNs. The reason is because IBP
starts to get loose as the number of sequences or layers increases. In
{\scshape DiffRNN}, we are dealing with two RNNs at the same time,
thus, the problem is twice harder. For example, 2-variable optimization
problems in {\scshape Popqorn} correspond to 4-variable optimization
problems in {\scshape DiffRNN}.

\section{Related work}
\ReluDiff{}~\cite{paulsen2020reludiff} is currently the only tool that can verify neural networks in the differential setting. However, unlike our approach, \ReluDiff{} does not solve the many challenges that are unique to RNNs. More generally, our work falls into the category of techniques for improving safety, security, and reliability in deep learning. Along this line, there has been a significant amount of research that we can classify into two broad categories: (1) techniques for discovering misbehaviors, and (2) techniques for proving the absence of misbehaviors, like \DiffRNN{}. We review a representative set of these works here.

Techniques along the first line are often geared towards finding \emph{adversarial examples}~\cite{szegedy2013intriguing, KurakinGB17a}. There have been many works using machine learning techniques such as gradient-based optimization and even generative adversarial networks~\cite{GoodfellowSS15, NguyenYC15, XuQE16, Moosavi-Dezfooli16}. In addition, other techniques use white-box heuristics~\cite{MaLLZG18, odena2018tensorfuzz, PeiCYJ17, TianPJR18, ma2018deepgauge, SunWRHKK18} such as neuron coverage or various black-box techniques~\cite{xie2019deephunter, xie2019diffchaser, WickerHK18}. While useful for discovering misbehavior they do not guarantee the absence of misbehavior, which we do.

Techniques along the second line usually aim to \emph{prove the absence of} adversarial examples.
For example, many works have developed exact and complete techniques that are guaranteed to \emph{eventually} terminate with the correct result. They have used LP solvers~\cite{HuangKWW17, RuanHK18, CarliniW17, BastaniILVNC16, DvijothamSGMK18, Ehlers17}, built specialized solvers for neural networks~\cite{KatzBDJK17, KatzHIJLLSTWZDK19, GopinathKPB18}, or combined approximation techniques with refinement~\cite{SinghGPV19iclr, WangPWYJ18, WangPWYJ18nips}. 

Others have focused solely on approximation techniques~\cite{zhang2018efficient, GehrMDTCV18,lyu2019fastened, SinghGPV19, WengZCSHDBD18}, which often use abstract domains~\cite{CousotC77}, such as intervals~\cite{moore2009introduction}, zonotopes~\cite{ghorbal2009zonotope}, and polyhedra~\cite{CousotH78}. Only very recent works have attempted to verify RNNs~\cite{ko2019popqorn, jia2019certified, shi2020robustness}, but, as we have shown, they do not perform well in the differential setting.

In addition, these techniques have been integrated into the training process to produce more robust and verifiable networks~\cite{RaghunathanSL18, WongK18, madry2017towards, FischerBDGZV19, MirmanGV18}. We believe a similar approach could be taken to produce networks more amenable to differential verification. We leave this as future work.

\section{Conclusion}
We have presented {\scshape DiffRNN}, the first method for
differential verification of two closely related recurrent neural
networks. By reasoning about general nonlinear activiation functions, our work goes beyond previous approaches for differential verification such as {\scshape ReluDiff} (that used only \emph{RelU} activiations). More crucially, we show how we can extend our approach to a more general class of NNs, known as recurrent neural networks. {\scshape DiffRNN} leverages interval analysis
to directly and more accurately compute difference in the values of
neurons of the two networks from the input layer to output layer. At
each step, the soundness of the computed differences is validated
using a nonlinear \emph{delta-sat} SMT solver. Our experimental comparison of {\scshape
DiffRNN} with state-of-the-art verification tools such as {\scshape
Crown} and {\scshape Popqorn} show that the proposed method not only
is faster but also can verify significantly more properties.  
%

\bibliographystyle{unsrt}  
\bibliography{references}  

\clearpage
\newpage
\section*{Appendix}
\label{sec:appendix}
\subsection{Many-to-one vanilla RNN}
\begin{figure}[h]
\centering
\includegraphics[scale=0.6]{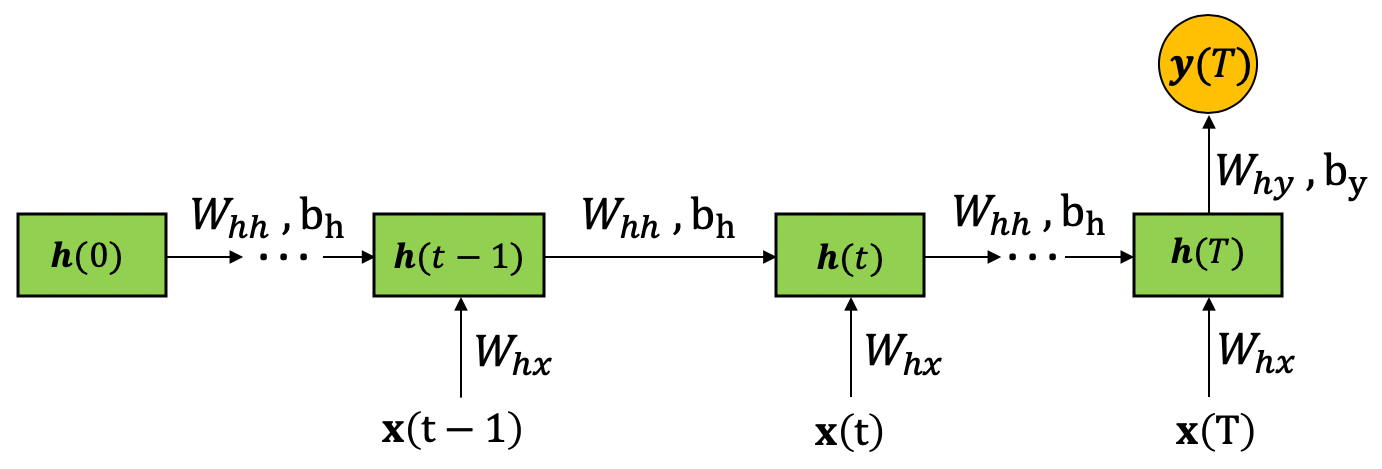}
\caption{A many-to-one Vanilla RNN}
\label{fig:vanilla_rnn}
\end{figure}

Fig.~\ref{fig:vanilla_rnn} describes the structure of a many-to-one
vanilla RNN. It consists of a single hidden layer, and at each time
step $t$ operates on input $\vInputSeq(t)$ and computes the output
$\vHid(t)$. At time $T$, it produces the output $\vOutputSeq(T)=
\whout\cdot\vHid(T)+ \biasout$.

\subsection{Algorithm for differential verification of LSTMs}
This appendix provides the algorithm for differential verification of
LSTMs (Algo.~\ref{alg:d_lstm}), and describes the nonlinear
transformer for LSTMs in Algo.~\ref{alg:non-linear_lstm}.
\begin{algorithm}[h]
\small
\caption{Differential verification of LSTMs. \label{alg:d_lstm}}
\SetKwProg{Fn}{Function}{:}{}

\KwIn{\textit{network f, network $f^\prime$, input region X}}
\KwOut{$\diffy$}
\vspace{0.2cm}
\textbf{Init:} Initialize $[L(\vHid(0)), U(\vHid(0))]$ and $[\underline{\diffh}(0), \overline{\diffh}(0)]$ to 0\;
\For{$t$ : 1 to $T$}{
{\color{teal} \tcp{affine transformer}}
{\color{teal} \tcp{symbolic values}}
Compute $[L(\igate_{\uin}(t)), U(\igate_{\uin}(t))]$;\\
Compute $[L(\fgate_{\uin}(t)), U(\fgate_{\uin}(t))]$;\\
Compute $[L(\ggate_{\uin}(t)), U(\ggate_{\uin}(t))]$;\\
Compute $[L(\ogate_{\uin}(t)), U(\ogate_{\uin}(t))]$;\\
{\color{teal} \tcp{concrete differences}}
Compute $[\underline{\diffi}(t), \overline{\diffi}(t)]$;\\
Compute $[\underline{\difff}(t), \overline{\difff}(t)]$;\\
Compute $[\underline{\diffg}(t), \overline{\diffg}(t)]$;\\
Compute $[\underline{\diffo}(t), \overline{\diffo}(t)]$;\\
{\color{teal} \tcp{nonlinear transformer (Algo.~\ref{alg:non-linear_lstm})}}
Compute $[L(\vHid(t)), U(\vHid(t))]$;{\color{teal} \tcp{\Popqorn{}}}
Compute $[\underline{\diffh}(t), \overline{\diffh}(t)]$;{\color{teal} \tcp{\DiffRNN{}}}

}
Compute $[\underline{\diffy}(T), \overline{\diffy}(T)]$; {\color{teal} \tcp{final difference interval}}
\end{algorithm}
\newpage

\begin{algorithm}[ht]
\small
\caption{Nonlinear Transformer for LSTMs. \label{alg:non-linear_lstm}}
\SetKwProg{Fn}{Function}{:}{}

\KwIn{ value intervals  $\igate_{\uin}(t)$, $\fgate_{\uin}(t)$, $\ggate_{\uin}(t)$, $\ogate_{\uin}(t)$, difference intervals $\diffi_{\uin}(t)$, $\difff_{\uin}(t)$, $\diffg_{\uin}(t)$, $\diffo_{\uin}(t)$, $adjust$=0.01}
\KwOut{value $[L(\vHid(t)), U(\vHid(t))]$, difference $[\underline{\diffh}(t),\overline{\diffh}(t)]$}
\vspace{0.2cm}

Compute $[L(\cstate(t)), U(\cstate(t))]$;{\color{teal} \tcp{\Popqorn{}}}
Compute $\overline{\diffc_1}(t) = \max[\diffc_1(t)]$;  {\color{teal} \tcp{global optimizer}}

{\color{teal} \tcp{dReal validation}}
$sat = dReal.CheckSat(\overline{\diffc_1}(t) < \diffc_1(t))$;\\
\While{sat == True}{
$\overline{\diffc_1}(t) = \overline{\diffc_1}(t) + adjust$;\\
$sat = dReal.CheckSat(\overline{\diffc_1}(t) < \diffc_1(t))$;\\
}
\vspace{0.2cm}

Compute $\underline{\diffc_1}(t) = \min[\diffc_1(t)]$;{\color{teal} \tcp{global optimizer}}

{\color{teal} \tcp{dReal validation}}
$sat = dReal.CheckSat(\underline{\diffc_1}(t) > \diffc_1(t))$;\\
\While{sat == True}{
$\underline{\diffc_1}(t) = \underline{\diffc_1}(t) - adjust$;\\
$sat = dReal.CheckSat(\underline{\diffc_1}(t) > \diffc_1(t))$;\\
}

\vspace{0.2cm}
{\color{teal} \tcp{Using global optimizer and dReal validation}}
Compute $[\underline{\diffc_2}(t), \overline{\diffc_2}(t)]$;\\

{\color{teal} \tcp{add up the computed bounds}}
 $[\underline{\diffc}(t), \overline{\diffc}(t)]=[\underline{\diffc_1}(t), \overline{\diffc_1}(t)]+[\underline{\diffc_2}(t), \overline{\diffc_2}(t)]$;\\

\vspace{0.2cm}
Compute $[L(\vHid(t)), U(\vHid(t))]$;{\color{teal} \tcp{\Popqorn{}}}
{\color{teal} \tcp{Using global optimizer and dReal validation}}
Compute $[\underline{\diffh}(t), \overline{\diffh}(t)]$;\\

\end{algorithm}
\clearpage

\end{document}